\documentclass{article}
\usepackage{microtype}
\usepackage{graphicx}
\usepackage{subfigure}
\usepackage{booktabs}
\usepackage{hyperref}
\usepackage{colortbl}

\usepackage[accepted]{icml2024}
\usepackage{bbding}
\usepackage{multirow}
\usepackage{caption}
\usepackage[font=small, labelfont=it]{caption}
\usepackage{diagbox}
\usepackage{supertabular}
\usepackage{amsmath}
\usepackage{amssymb}
\usepackage{mathtools}
\usepackage{amsthm}
\usepackage{enumitem}
\usepackage{xcolor}
\definecolor{ggreen}{RGB}{112, 173, 71}
\usepackage[capitalize,noabbrev]{cleveref}

\theoremstyle{plain}

\theoremstyle{definition}

\theoremstyle{remark}

\usepackage[textsize=tiny]{todonotes}

\icmltitlerunning{RoboMP2: A Robotic Multimodal Perception-Planning Framework with Multimodal Large Language Models}

\begin{document}
\twocolumn[
\icmltitle{RoboMP$^2$: A Robotic Multimodal Perception-Planning Framework with Multimodal Large Language Models}

\icmlsetsymbol{equal}{$\dag$}

\begin{icmlauthorlist}
\icmlauthor{Qi Lv}{yyy,comp,wanda}
\icmlauthor{Hao Li}{yyy}
\icmlauthor{Xiang Deng}{equal,yyy}
\icmlauthor{Rui Shao}{yyy}
\icmlauthor{Michael Yu Wang}{comp}
\icmlauthor{Liqiang Nie}{equal,yyy}
\newline
\end{icmlauthorlist}
\centering{\href{https://aopolin-lv.github.io/RoboMP2.github.io/}{\textcolor{magenta}{\texttt{RoboMP2.github.io}}}}

\icmlaffiliation{yyy}{School of Computer Science and\newline Technology, Harbin Institute of Technology (Shenzhen)}
\icmlaffiliation{comp}{School of Engineering, Great Bay University}
\icmlaffiliation{wanda}{School of Computing and Information Technology, Great Bay University}

\icmlcorrespondingauthor{Xiang Deng}{dengxiang@hit.edu.cn}
\icmlcorrespondingauthor{Liqiang Nie}{nieliqiang@gmail.com}

\icmlkeywords{Machine Learning, ICML}

\vskip 0.3in
]
\printAffiliationsAndNotice{} 

\begin{abstract}
Multimodal Large Language Models (MLLMs) have shown impressive reasoning abilities and general intelligence in various domains. It inspires researchers to train end-to-end MLLMs or utilize large models to generate policies with human-selected prompts for embodied agents. 
However, these methods exhibit limited generalization capabilities on unseen tasks or scenarios, and overlook the multimodal environment information which is critical for robots to make decisions. 
In this paper, we introduce a novel \textbf{Robo}tic \textbf{M}ultimodal \textbf{P}erception-\textbf{P}lanning (\textbf{RoboMP$^2$}) framework for robotic manipulation which consists of a Goal-Conditioned Multimodal Preceptor (GCMP) and a Retrieval-Augmented Multimodal Planner (RAMP). 
Specially, GCMP captures environment states by employing a tailored MLLMs for embodied agents with the abilities of semantic reasoning and localization. 
RAMP utilizes coarse-to-fine retrieval method to find the $k$ most-relevant policies as in-context demonstrations to enhance the planner. 
Extensive experiments demonstrate the superiority of RoboMP$^2$ on both VIMA benchmark and real-world tasks, with around 10\% improvement over the baselines.
\end{abstract}

\section{Introduction}
Multimodal Large Language Models (MLLMs)~\cite{GPT4v, Llava-1.5} have exhibited remarkable intelligence in various vision and language tasks.
Inspired by the impressive performance, many efforts~\cite{SayCan, Monologue} have been made on using large models to empower embodied agents with human-like intelligence.
These approaches mainly use large models to train end-to-end policies~\cite{Palm-e, RT-1, RT-2} or generate plans with human-selected prompts~\cite{CaP, Voxposer}.
However, end-to-end policies and human-selected prompts lack generalization and flexibility on unseen tasks.

\begin{figure}[tbp]
    \centering
    \includegraphics[width=\linewidth]{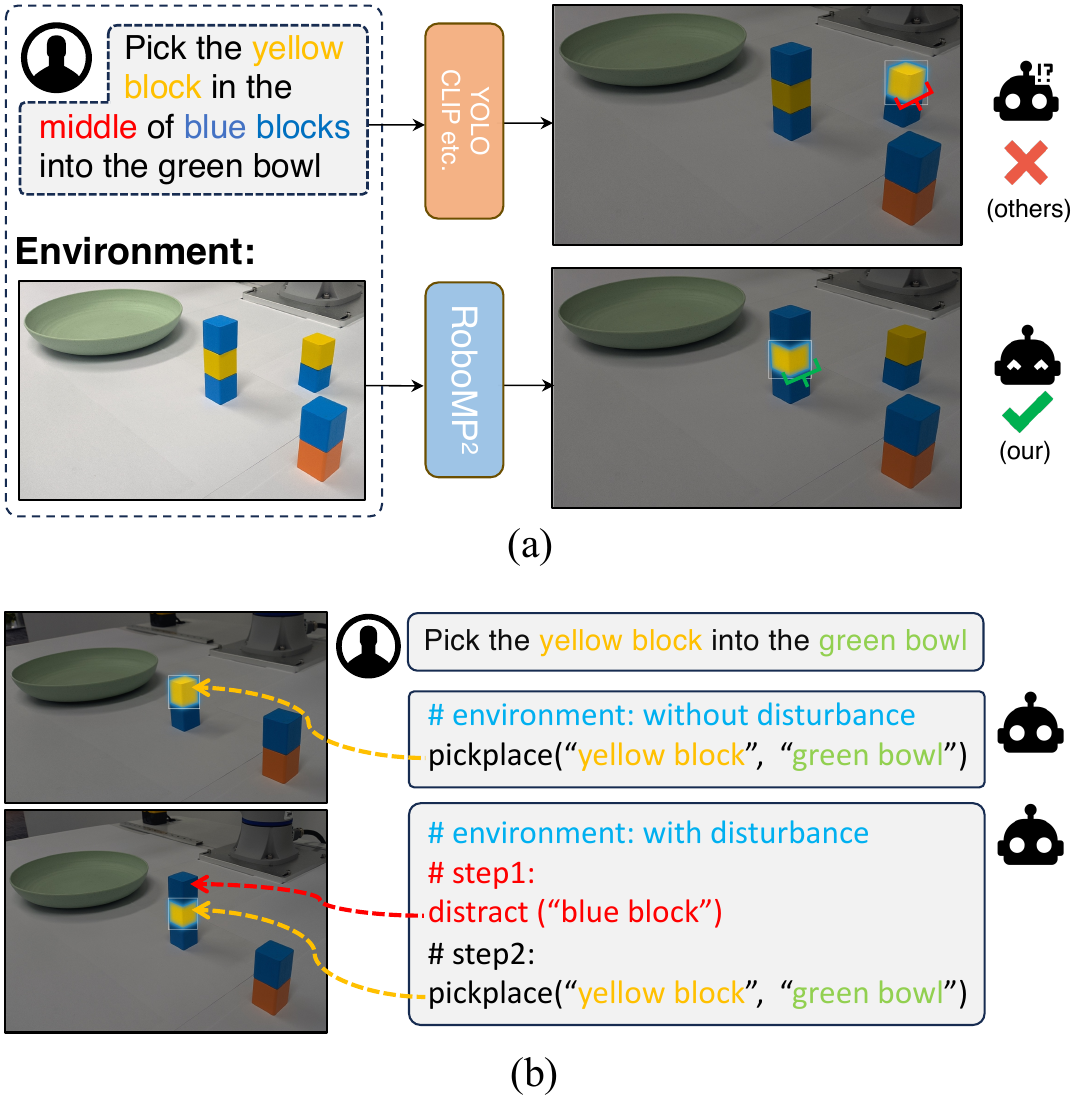}
    \vspace{-14pt}
    \caption{(a) The detection results of the yellow block with the complex spatial reference using different methods; (b) Different plans for different environment, even if the same instruction.}
    \label{fig:intro_1}
    \vspace{-12pt}
\end{figure}
Environment perception and task planning are two fundamental components for embodied agents to complete a task~\cite{SayCan, Openended}. 
Most of the existing work typically employs mainstream vision models as environment perceptors, such as YOLOv5~\cite{Yolo} and CLIP~\cite{CLIP}).
These models work well in simple scenarios where the categories of the objects are pre-defined or the relationships among objects are easy to be captured.
However, they lack the capability to identify and locate objects in unseen scenarios or objects with intricate spatial relationships.
For instance, as illustrated in \textit{Figure}~\ref{fig:intro_1}(a), the task entails picking up the ``\textcolor[RGB]{255,192,0}{yellow block} in the middle of the \textcolor{blue}{blue blocks}''. 
The existing perceptors struggle to accurately identify and locate the specified yellow block since these models cannot understand the semantic information of the complex referring expression.
Therefore, it necessitates the development of robot perceptors with multimodal understanding and reasoning capabilities. 
In light of this, we turn to MLLMs as environment perceptors since they have shown remarkable semantic understanding and vision perception abilities in various tasks~\cite{Llava-1.5}.
We accordingly propose a MLLM-based Goal-Conditioned Multimodal Perceptor (GCMP) with an advanced ability to detect objects with a given semantic-complex reference goal.
As shown in \textit{Figure}~\ref{fig:intro_1}(a), GCMP enables robots to accurately identify and locate referred objects while the per-\newline ceptors adopted in the existing embodied frameworks fail.

In addition to the multimodal environment perceptor, the planning for subsequent execution is also critical.
The existing policies mainly include end-to-end models~\cite{RT-1} and prompt-based approaches~\cite{Voxposer}.
The end-to-end policy integrates the perception and planning into a single model, thus requiring closed-loop robot data.
However, in the real world, the closed-loop data is very limited due to the expensive human labor for collection.
Consequently, these models are proven to overfitting~\cite{SayCan, vima} the data-collection scenario and show limited generalization in unseen environments or on new tasks.
On the other hand, current prompt-based methods rely on manually designed and selected prompt templates to prompt LLMs to generate plans for a given task.
They inherently lack generalization~\cite{HowToPrompt, Ins2Act} in diversities of tasks that are highly different from the tasks given in the demonstrations of the prompt templates.
To cover different types of tasks, these methods adopt large amounts of templates, which results in in-context attention distraction.
\citet{GPT3makes} have also demonstrated that redundant in-context examples fail to offer effective guidance, which is detrimental to the ability of the model~\cite{GPT3makes}.
More importantly, these approaches~\cite{Monologue, ChatGPTForRobotics} typically generate plans based solely on the text instruction, overlooking critical environmental information.
For example, as shown in \textit{Figure}~\ref{fig:intro_1}(b), even for the same instruction, different environments lead to different plans.
To address these issues, we propose a retrieval-augmented method which prompts MLLMs to generate plans via adaptively choosing the most relevant policies as demonstrations.

In this paper, we aim to explore how to sufficiently leverage both multimodal information in an environment and the general intelligence in large models to enhance the perception and reasoning capabilities of embodied robots.
We propose a novel \textbf{Robo}tic \textbf{M}ultimodal \textbf{P}erception-\textbf{P}lanning framework (\textbf{RoboMP$^2$}) which is composed of a Goal-Conditioned Multimodal Perceptor (GCMP) and a Retrieval-Augmented Multimodal Planner (RAMP).
Specifically, to endow the embodied model with semantic reasoning and localization capabilities, we propose GCMP to comprehensively perceive environmental information through the integration of a tailored MLLM. Meanwhile, to enhance the generalization of policy planning, we devise RAMP to adaptively find the $k$ most-relevant policies as in-context demonstrations with dedicated coarse retriever, fine reranker.
Our main contributions are as followed:
\begin{itemize}
    \item Different from the existing robot perceptors that can only identify objects with pre-defined classes or simple references, we introduce a taiored MLLM as the environment perceptor, i.e., GCMP, which owns the comprehension abilities to perceive targeted objects with complex references.
    \item Different from the existing code planners that simply generate code based solely on a text instruction with manually selected templates, we propose RAMP that integrates multimodal environment information into the code generation process, and develops a retrieval-augment strategy to mitigate the interference of redundant in-context examples.
    \item Extensive experiments show that our proposed RoboMP$^2$, which is composed of GCMP and RAMP, outperforms the baselines by around 10\% on both VIMABench and real-world tasks.
\end{itemize}

\begin{figure*}[h!]
    \begin{center}
    \includegraphics[width=0.9\linewidth]{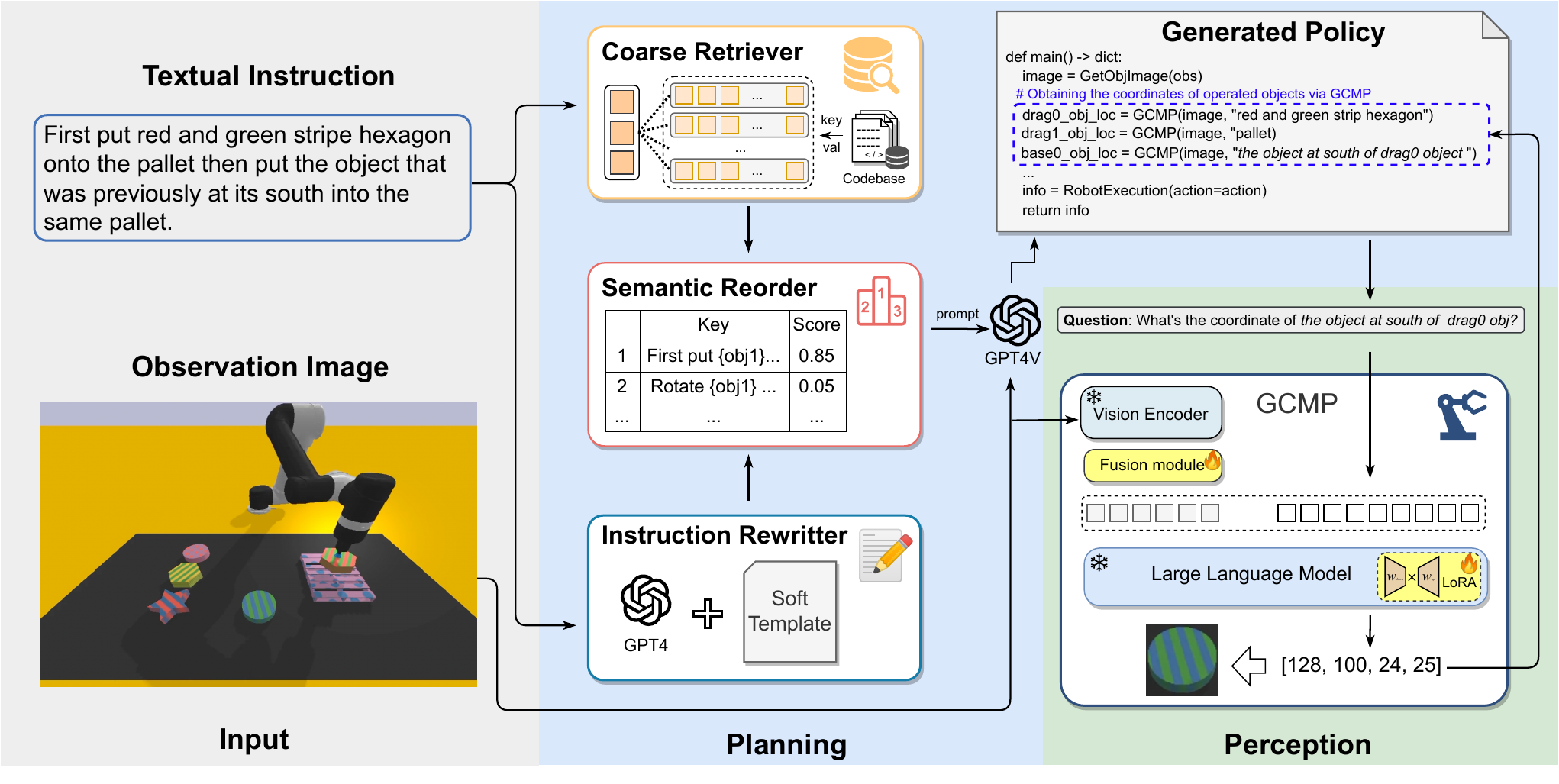}
    \caption{Overview of our proposed \textbf{RoboMP$^2$} framework. The three parts in grey/blue/green represent the input data, planning and perception, respectively. The modules highlighted are trainable, including fusion module and LoRA.}
    \label{fig:RoboMPE_framework}
    \vspace{-8pt}
    \end{center}
\end{figure*}
\section{Related Work}
\textbf{MLLMs for Robotic Manipulation.}
Robotic manipulation aims to complete a specific task by interacting with the environment.
In recent, imitation learning~\cite{RT-1, RT-2, DiffusionPolicy} has achieved a great success in robotic learning.
However, due to the task complexity, it needs large amounts of data to train a robot agent to achieve strong generalization capability~\cite{RH20T, OpenX-Embodiment, PerAct}, which is time-consuming and computational-unfriendly.
Therefore, many efforts~\cite{Ins2Act, Voxposer, RoboTool} have been made on prompting LLMs to generate policies in a zero-shot manner to control robot agent.
\citet{CaP} proposed using programs as the policy to control robots by generating calls to perception APIs and control APIs.
It has been proven that the prompt is crucial when using LLMs to generate text in a zero-shot manner~\cite{GPT3makes}.
Thus, many approaches~\cite{HowToPrompt, ChatGPTForRobotics, ProgPrompt} explored the construction of the text prompt, including descriptions of the task details, robotic API libraries, environment feedback~\cite{Openended}.
One constraint of previous studies is that they merely utilized the textual information to prompt LLMs, ignoring the significant multimodal information of the environment.
Moreover, it would confuse the model if there are too many prompt templates in-context examples, resulting in a wrong execution plan.
Thus, in order to alleviate these problems, we introduce a retrieval-augmented multimodal planner which sufficiently leverages the multimodal information and most relevant demonstrations for task planning.

\textbf{Retrieval Augmented Generation with MLLMs.}
Retrieval augmented generation (RAG) was first introduced to serve as more informative inputs to unlesh the extensive knowledge of LLMs~\cite{RetrievalAugmented, GPT3makes}.
Due to its effectiveness, it was subsequently introduced to the multimodal domain.
It assists models generate answers by retrieving contents related to the original input as supplement context.
RETRO~\cite{Retro} introduced additional retrieval encoders to train a GPT model from scratch~\cite{gpt2}.
Atlas~\cite{Atlas} incorporated similar retrieval encoders to continually finetune T5 model~\cite{t5}.
In addition, KNN-based retrieval methods~\cite{knn-retrieve} and cross modal semantic-based retrieval methods~\cite{clip-retrieve} are commonly used to gather information from different modalities, aiming to generate more satisfied text by providing evidence.
Considering the plans for robotic manipulation tasks are predominantly involved in the executed actions and target objects, we incorporate a task rewriting module. 
This module is introduced to extract essential textual information from task instructions. 
By integrating this module with a coarse-to-fine retrieval method, we adaptively select policies that exhibit semantic similarity as demonstrations, aiming to unveil the inherent capabilities of MLLMs.

\textbf{Task and Motion Planning.}
It~\cite{Garrett2020IntegratedTA, silver2023predicate, 5980391} selects the sequence of high-level actions~\cite{FIKES1971189, Nau1999SHOPSH} that the robot should take, the hybrid parameter values that determine how the action is performed, and the low-level motions~\cite{LaValle_2006} that safely execute the action. Traditionally, these approaches build on research through optimization~\cite{10.5555/2832415.2832517} or symbolic reasoning~\cite{kumar2022overcoming}, but more recently machine learning has been applied to aspects of the problem via learned representations~\cite{silver2023predicate, NEURIPS2019_5c48ff18, xu2018neural, xu2019regression, shah2021value}, learned task primitives, and more. Some works utilize language for planning and grounding~\cite{huang2022language, Monologue, SayCan, 5453186, 10.5555/2900423.2900661}. Others have approached the problem through hierarchical learning~\cite{nair2019hierarchical, xia2020relmogen, shah2021value, hafner2022deep}. In this work, we leverage MLLMs and their semantic knowledge to find feasible plans.

\section{Framework}
In this section, we present a robotic multimodal perception-planning framework (RoboMP$^2$) that utilizes a Goal-Conditioned Multimodal Perceptor (GCMP) and a Retrieval-Augmented Multimodal Planner (RAMP) to enhance the perception and planning abilities of embdied agents.
The framework of RoboMP$^2$ is shown in \textit{Figure}~\ref{fig:RoboMPE_framework}.

\subsection{Goal-Conditioned Multimodal Perceptor}\label{sec:LCP}
\subsubsection{Environment Perception for Manipulation}\label{sec:LCP_complex_refer}

Robotic manipulation involves various types of tasks, imposing different requirements on the perceptor of the robot.
For example, when a task entails picking up ``an apple", the robot perceptor needs to have the ability to identify and locate the apple for robot execution.
While the perceptors in the existing studies can work well in recognizing this kind of simply referred objects, e.g., pre-defined classes of objects, they struggle to identify semantic-complex referred objects.
For instance, when the task involves picking up ``the two green apples on the left of the yellow cup", the existing perceptors fail, since they cannot understand the instruction semantics or the spatial relationships among different objects.
In practice, these complex references about objects widely exist in different tasks.
We provide three types of commonly used reference expressions that can hardly be handled by the existing perceptors, i.e., referring objects based on attributes, spatial relationships, and knowledge reasoning:

\begin{figure*}[h!]
    \centering
    \includegraphics[width=0.99\linewidth]{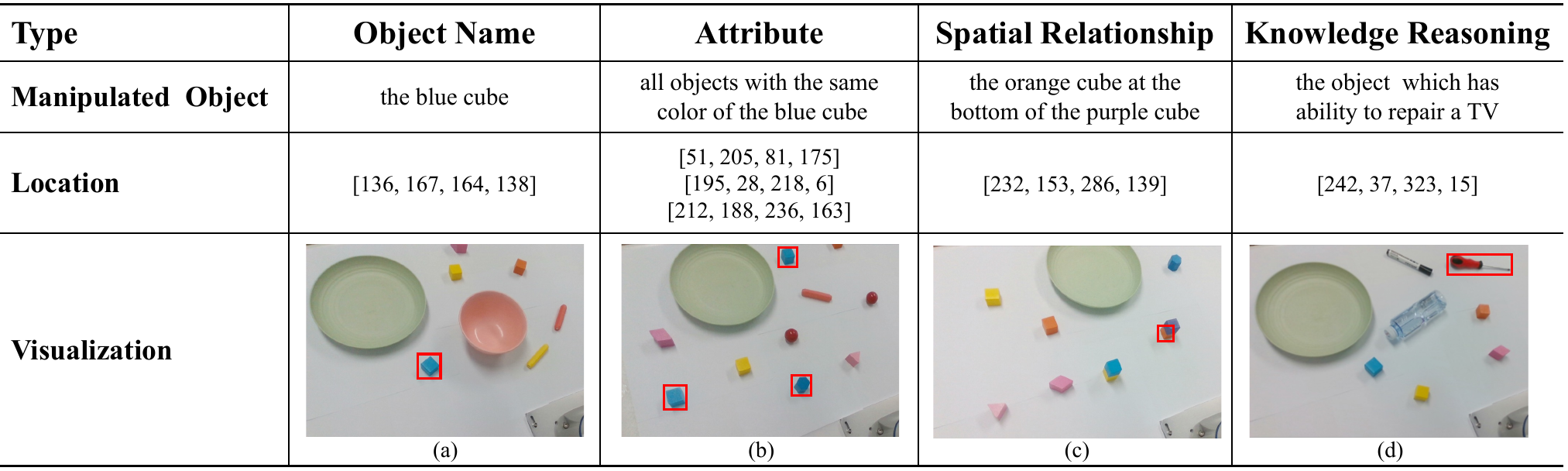}
    \captionof{figure}{Examples of manipulated objects with four different referential relationships types.}
    \label{fig:referential_type}
\end{figure*}
\textbf{(1) Object Perception Based on Attributes:}
Many tasks require robots to manipulate objects with specific attributes. As shown in \textit{Figure}~\ref{fig:referential_type}(b), the robot is asked to manipulate ``all the objects with the same color of the blue cube''.
It involves one-to-many perception, where a single attribute pertains to many objects.
This kind of manipulation requires the perceptor to have a powerful attribute-aware perception ability for not leaving out any of the referred objects.

\textbf{(2) Object Perception Based on Spatial Relationships:}
Robots often receive instructions to manipulate objects at specific positions. 
For example, in \textit{Figure}~\ref{fig:referential_type}(c), there are two orange blocks, and the instruction is to manipulate ``the orange block at the bottom of the purple block''. 
This requires the perceptor to have spatial-aware perception capabilities.

\textbf{(3) Object Perception Based on Knowledge Reasoning:}
Robot manipulation also face situations where knowledge reasoning is needed to determine the operational target based on instructions. 
As illustrated in \textit{Figure}~\ref{fig:referential_type}(d), supposing the television stops working, people ask the robot for handing him an object capable of repairing the television. 
It requires the robot to have high-level understanding and reasoning abilities.

The existing studies~\cite{CaP, Ins2Act} which employ Yolov5 or CLIP can locate objects with the basic object name or simple semantics, such as the example in \textit{Figure}~\ref{fig:referential_type}(a). However, they lack the ability of perceiving objects with complex referential expressions, such as the the above three kinds of expressions.
To address this issue, we built a Goal-Conditioned Multimodal Perceptor (GCMP) upon a MLLM that have both semantic understanding and vision perception abilities.

\subsubsection{Training of Multimodal Perceptor}
\begin{figure}[tbp]
    \centering
    \includegraphics[width=0.95\linewidth]{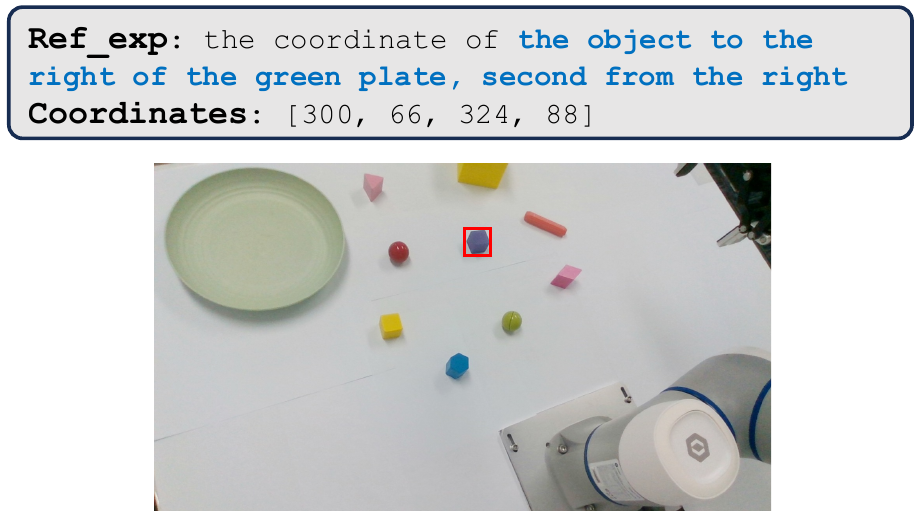}
    \caption{An example of training data for the GCMP.}
    \vspace{-10pt}
    \label{fig:ift_sample}
\end{figure}

To learn a tailored multimodal perceptor for embodied agents, we formulate robotic data into instruction-tuning format, namely \{\texttt{image}, \texttt{ref\_exp}, \texttt{coordinates}\} triplets, where the input consists of an image and a referential expression (ref\_exp), and the output is the corresponding coordinates for manipulation.
An example is shown in \textit{Figure}~\ref{fig:ift_sample}.
For model architecture, we adopt the ViT~\cite{vit} and flan-t5-xl~\cite{flan-t5} as our embodied vision encoder and language encoder.
Then, we connect these two encoders through a MLP layer.
In addition, we introduce a LoRA module~\cite{lora} to tune the LLM efficiently and effectively.

During instruction tuning, we employ the natural language formulation of all question-answer pairs as the input $(\mathbf{x}^\mathrm{I}, \mathbf{x}^\mathrm{T})$ and output $\mathbf{y}^\mathrm{T}$.
The answer is also uniformly formulated as natural language and we adopt the autoregressive training paradigm to tune the model.
The learning objective is as followed:
\begin{gather}
    \mathcal{L} = -\sum_k \log P(y^\mathrm{T}_{k+1}|y^\mathrm{T}_{i<k},(\mathbf{x}^\mathrm{I}, \mathbf{x}^\mathrm{T}); \theta),
\end{gather}
where the $\theta$ denotes the learnable weights of the model and $\mathbf{x}^\mathrm{I}$ denotes the environment image.

\subsection{Retrieval-Augmented Multimodal Planner}
Planning is critical for embodied agents to complete tasks.
Exiting approaches typically prompt LLMs to generate plans according to a textual task instruction and a manually selected prompt template.
However, they suffer from two issues: 
(1) utilizing human selected templates for a given instruction while hardly generalizing to new tasks;
(2) only using text information while ignoring multimodal environment information.
To address these challenges, we introduce RAMP which is composed of a coarse retriever, a instruction and a fine reranker to adaptively find the $k$ most-relevant policies as in-context demonstrations, thereby boosting the generalization of the policy planning.

\subsubsection{Coarse Retriever}
To cover diversities of tasks, a prompt typically includes many in-context examples as demonstrations to prompt LLMs to generate task plans.
However, this also leads to attention diffusion across these examples, which hurts the quality of the generated plan.
To address this issue, we propose a coarse retriever to find the most relevant policies from a codebase.
Specifically, we initially assemble a codebase comprising over 50 programs gathered from 10 diverse sources~\cite{HowToPrompt, RoboTool, CaP}.
Given the textual query $\mathbf{x}^\mathrm{que}$ of a task instruction, the coarse retriever aims to recall $K$ programs $\{\mathbf{x}_i^\mathcal{C}\}_{i=1}^{K}$ from codebase $\mathcal{C}$, where $N$ represents the number of target candidates.
We adopt the two-tower architecture $f\left( \mathbf{\cdot} \right)$ which extracts the information of $\mathbf{x}^\mathrm{que}$ and $\mathbf{x}_i^\mathcal{C}$ separately.
The score between task instruction query $\mathbf{x}^\mathrm{que}$ and candidate $\mathbf{x}_i^\mathcal{C}$ is as followed:
\begin{gather}
    \mathbf{V}^{\mathrm{que}} = f(\mathbf{x}^{\mathrm{que}}), \ \ 
    \mathbf{V}_i^{\mathcal{C}} = f(\mathbf{x}_i^{\mathcal{C}}), \\
    \mathrm{Score}_i^\mathrm{CR} = \mathbf{V}^{\mathrm{que}} \circ \mathbf{V}_i^{\mathcal{C}},
\end{gather}
where $\circ$ represents the dot product operation. 

We adopt TF-IDF~\cite{TF-IDF} to compute the similarity.
Meanwhile, extensive experiments are conducted to compare this metric and other two classic metrics, i.e., BM25~\cite{BM25} and SentenceBERT~\cite{SBERT}.
The details of similarity computations and the experimental comparison can be seen in Appendix~\ref{appendix:dense_retriever} and Section~\ref{sec:exp_retreiver}.

\subsubsection{Fine Reranker}
Despite relevant code snippets could be recalled through the coarse retriever, there remains an issue of the order among candidates. 
In the planning generation phase, large models are highly sensitive to the sequence of demonstrations.
Consequently, we introduce a rewriting module and a reranker module respectively to extract its core of the task instruction and order these relevant demonstrations.
Finally, we only use the $k$ most relevant programs as the in-context examples.

\paragraph{Instruction Rewriter.}
The robotic policy is mainly related to the action type and manipulation objects.
Thus, we introduce a instruction rewriting module to eliminate distracting expressions from the task description, obtaining its core robotic operation instructions.
Specifically, we combine the raw textual task query $\mathbf{x}^{\mathrm{que}}$ and a soft prompt template to rewrite $\hat{\mathbf{x}}^{\mathrm{que}}$.
The formula is as followed:
\begin{gather}
    \hat{\mathbf{x}}^{\mathrm{que}} = \mathrm{Rewritter}(\mathbf{x}^{\mathrm{que}}, T_{\mathrm{soft}}),
\end{gather}
where we use the GPT4 as the rewritter and $T_{\mathrm{soft}}$ denotes the soft prompt template which consists of \texttt{Scene}, \texttt{Description}, \texttt{Example} and \texttt{Instruction}.
The first two primarily provide premises and requirements for the robot scenarios, while the latter two consist of concrete examples and the corresponding task instructions.
More details can be seen in Appendix~\ref{appendix:rewriting_prompt}.

\paragraph{Semantic Reorder.}
In the plan generation stage, the order of the examples is crucial.
Due to the design of positional encoding in the transformer architecture, researchers~\cite{order_in_context, Learning_in_context} have found that model generation tends to primarily focus on the content at the beginning and the end of the paragraph.
Therefore, we introduce a reorder module with the aim of sorting the retrieved candidate code snippets$\{\mathbf{x}_i^\mathcal{C}\}_{i=1}^{K}$.
A straightforward solution is to simply concatenate it at the beginning of the whole candidate.
\begin{gather}
    \mathbf{H}_i = \mathrm{SentenceBERT}([\hat{\mathbf{x}}^\mathrm{que}, \mathbf{x}^\mathcal{C}_{i}]), \\ 
    \mathrm{Score}_i^\mathrm{FR} = \mathrm{Sigmoid}(\mathbf{H}_i).
\end{gather}

In the end, the coarse-grained ranking scores and fine-grained ranking scores would be applied for the comprehensive ranking as followed:
\begin{align}
    \mathrm{Score}_i &= \lambda \cdot \mathrm{Score}_i^\mathrm{CR} + (1-\lambda) \cdot \mathrm{Score}_i^\mathrm{FR}, \\
    &\propto p(x_\mathrm{[cls]}|\mathbf{x}^\mathrm{que}, \hat{\mathbf{x}}^\mathrm{que}, \mathbf{x}^\mathcal{C}_{i})\label{eq:propto}.
\end{align}
Equation~\ref{eq:propto} illustrates the relationship between the $\mathrm{Score}_i$ and $(\mathbf{x}^\mathcal{C}, \mathbf{x}^\mathrm{que}_i, \hat{\mathbf{x}}^\mathrm{que}_i)$, when SentenceBERT is employed for both the coarse retriever and the fine reranker.

\begin{figure*}[h]
  \centering
  \begin{minipage}{.61\textwidth}
    \centering
    \resizebox{0.99\linewidth}{!}{
    \begin{tabular}{llllll}
    \toprule[1pt]
          & L1 & L2 & L3 & L4 & Avg. \\
    \midrule
    \textit{End-to-end models} &       &       &       &       &  \\
    \midrule
    Gato$^\dag$ & 58.1  & 53.2  & 43.5  & 12.4  & 41.8  \\
    Flamingo$^\dag$ & 47.5  & 46.0  & 40.8  & 12.1  & 36.6  \\
    VIMA$^\dag$ & 81.6  & 81.5  & 79.0  & 48.9  & 72.7  \\
    RT-2 & 72.8 & 70.3  & 66.8  & 47.0  & 64.2  \\
    \midrule
    \textit{MLLM Planners} &       &       &       &       &  \\
    \midrule
    CaP & 71.2  & 70.0  & 42.8  & 44.7  & 57.2  \\
    VisualProg & 49.7  & 47.7  & 69.9  & 52.2 & 54.9  \\
    I2A$^\dag$ & 77.0  & 76.2  & 66.6  & 49.8  & 65.0  \\
    RoboMP$^2$ & \textbf{89.0} \textcolor{ggreen}{(+7.4)}  & \textbf{85.9 }\textcolor{ggreen}{(+4.4)}  & \textbf{86.8} \textcolor{ggreen}{(+7.8)}   & \textbf{68.0} \textcolor{ggreen}{(+19.1)}  & \textbf{82.4} \textcolor{ggreen}{(+9.7)}  \\
    \bottomrule
    \end{tabular}}
    \captionof{table}{The results on the VIMA Benchmark. $^\dag$ denotes the cited result.}
    \label{tab:vima_result}
  \end{minipage}
  \begin{minipage}{.35\textwidth}
    \centering
    \resizebox{0.98\linewidth}{!}{
    \begin{tabular}{lrr}
    \toprule
    Task Name & I2A & RoboMP$^2$ \\
    \toprule
    \multicolumn{3}{l}{\textit{Basic Task}} \\
    \midrule
    Visual Manipulation & 85.0    & 90.0 \\
    Same Shape & 45.0     & 90.0 \\
    Same Color & 40.0     & 90.0 \\
    \midrule
    \multicolumn{3}{l}{\textit{Challenging Task}} \\
    \midrule
    Manipulate Old Neighbor & 0.0     & 70.0 \\
    Pick in Order then Restore & 65.0     & 80.0 \\
    Interfering Manipulation &0.0 & 55.0 \\
    \midrule
    Avg. Success Ratio (\%) & 39.2    & \textbf{79.2} \\
    \bottomrule
    \end{tabular}}%
    \captionof{table}{The results on the real-world tasks.}
    \label{tab:real_scene_result}
  \end{minipage}
  \label{fig:side_by_side_tables}
\end{figure*}
\subsubsection{Multimodal Generation Module}
After adaptively selecting $k$ most relevant examples $\{\mathbf{x}_i^\mathcal{C}\}_{i=1}^{k}$ 
through the coarse-to-fine retrieval method, we combine them with a template to construct the complete prompt, including the third-party libraries, function definitions, and the task instruction.
It should be noted that we arrange the examples in reverse order based on relevance, meaning the most-relevant examples are placed towards the end.
Compared with the pure textual generator, the multimodal information of current environment is crucial.
Hence we use the GPT4V as the multimodal generator with the input $\{\mathbf{x}^\mathrm{que}, \mathbf{x}^\mathrm{I}, \mathbf{x}_i^\mathcal{C}\}_{i=1}^{k}$ and full prompt template.
Examples of full template are presented in Appendix~\ref{full_template}.

\section{Experiment}
\subsection{Data and Evaluation Metrics}
We employ VIMA~\cite{vima} as the test benchmark which encompasses 17 tasks ranging from L1-level to L4-level difficulty. 
In the zero-shot real-world experiment, we construct 7 tasks including 4 sim-to-real tasks and 3 generalization tasks.
We measure the performance of different methods through the Success Rate (SR) metrics.
Please refer to Appendix~\ref{appendix:dataset} for more details.

\subsection{Baselines}
We compare our method with the following two types of baselines:

\textbf{End-to-end models}:
(1) \textbf{Gato}~\cite{gato} introduces a decoder-only architecture which is prompted with the observation and action subsequence;
(2) \textbf{Flamingo}~\cite{Flamingo} embeds a variable number of prompt images into a fixed number of tokens as Perceiver and an additional robot action heads;
(3) \textbf{VIMA}~\cite{vima} leverages the multimodal prompt and x-attention block to fuse the instruction and observation information;
(4) \textbf{RT-2}~\cite{vima} trains a large robotic model by stacking multi-transformer decoder blocks via autoregressive generation.

\textbf{Prompt-based methods}:
(5) \textbf{VisProg}~\cite{Ins2Act} generates python-like modular programs, which are then executed to get both the solution and a comprehensive and interpretable rationale.
(6) \textbf{CaP}~\cite{CaP} generates the code as policies to manipulate the robot given the textual instruction.
(7) \textbf{I2A}~\cite{Ins2Act} follows the code policy style while introducing the off-the-shelf SAM~\cite{SAM} and CLIP~\cite{CLIP} to obtain coordinates of target objects.

\subsection{Implementation Details}
For GCMP in our proposed RoboMP$^2$, we adopt the \texttt{EVA-CLIP/g}~\cite{eva} as the visual encoder of our perceptor.
Regarding the LLM, we investigate the encoder-decoder architecture model \texttt{flan-t5-xl}~\cite{flan-t5}.
During training, the vision encoder and the LLM are frozen, only learning the weight of the fusion module and the LoRA module.
We set the epoch to 10, the batch size to 128, the learning rates of the fusion module and LoRA module to 3e-5 and 1e-4, respectively.
We adopt the AdamW optimizer and the cosine decay learning schedule.
The overall training time is around 24 hours on a 8*A100-80G-SXM4 platform.

For RAMP in the proposed RoboMP$^2$, we adopt the GPT4/GPT3.5 and GPT4V as the text-only rewritter and the multimodal generator.
The number of the most-relevant examples during coarse retriever and fine reranker is set to 5 and 2, respectively.

\subsection{Results}
\paragraph{Experimental Results on VIMABench.}

The task difficulty of VIMABench ranges from L1 level to L4 level, with tasks at L1-L3 levels being seen, while tasks at L4 level are unseen.
As shown in \textit{Table}~\ref{tab:vima_result}, our RoboMP$^2$ outperforms other methods by a large extent, surpassing the VIMA baseline around 10\% on the average performance of success ratio.
This demonstrates the effectiveness of RoboMP$^2$ for robotic manipulation.

Compared with the other baselines, the average performance improvements of RoboMP$^2$ range from 4\% to 8\% on seen tasks (L1-L3), while a comprehensive improvement of approximately 20\% on unseen tasks (L4).
The reason is that end-to-end models tend to overfit to the training tasks due to the small amount of closed-loop data, resulting in limited generalization abilities on unseen tasks.
The other prompted-based methods use fixed human-selected in-context examples in the prompt, thus lacking adaptability on diverse unseen tasks.
In contrast, the proposed RoboMP$^2$ shows a much better generalization ability on unseen tasks since RAMP in RoboMP$^2$ can adaptively find the most relevant policies as their demonstrations to prompt the generator according to the task.

\begin{table*}[h!]
  \centering
  \resizebox{0.73\linewidth}{!}{
    \begin{tabular}{lccccccc}
    \toprule
          & \multicolumn{1}{p{5em}}{\centering Visual\newline{}manipulate} & \multirow{2}{*}{\centering Rotate} & \multirow{2}{*}{\centering Rearrange} & \multicolumn{1}{p{5em}}{\centering  Rearrange then  restore} & \multicolumn{1}{p{5em}}{\centering Scene \newline{}understanding} & \multicolumn{1}{p{7em}}{\centering Pick in order\newline{} then restore} & \multirow{2}{*}{Avg.} \\
    \midrule
    I2A   & 63.5  & 3.4   & 0.0   & 0.0   & 2.9   & 8.4 & 13.0 \\
    RoboMP$^2$ & \textbf{92.0}  & \textbf{52.0}  & \textbf{84.0}  & \textbf{88.0}  & \textbf{60.0}  & \textbf{28.0} & \textbf{67.3} \\
    \bottomrule
    \end{tabular}}
  \caption{The results on the enriched test set.}
  \vspace{-12pt}
  \label{tab:exp_enrich}
\end{table*}

\paragraph{Real-world Experimental Results.}
We conduct experiments on 6 real-world tasks, comparing the performance of RoboMP$^2$ with that of I2A. This is due to that both of them can be transferred from simulation to reality, without requiring to collect data for additional training.
As illustrated in \textit{Table}~\ref{tab:real_scene_result}, our RoboMP$^2$ outperforms I2A by 40\% in terms of the overall average success ratio.
It can be seen that I2A performs well in the task \texttt{visual manipulation}.
However, it struggles to complete the manipulation tasks of objects with specific attributes such as \texttt{same shape}, let alone challenging tasks.
In comparison, RoboMP$^2$ shows consistent improvements across these two tasks.

It is noted that, on the tasks requiring to recognize objects with the same attribute or a specific relationship such as \texttt{manipulate old neighbor}, the performance of I2A drops significantly, even to zero.
The reason is that the SAM+CLIP perceptor in I2A is unable to simultaneously identify multiple instances and is hardly handle the detection of objects with complex relationships.
Benefiting from our proposed Goal-Conditioned Multimodal Perceptor (GCMP), RoboMP$^2$ demonstrates a strong capability in dealing with such challenges.
\begin{figure*}[h!]
    \includegraphics[width=\linewidth]{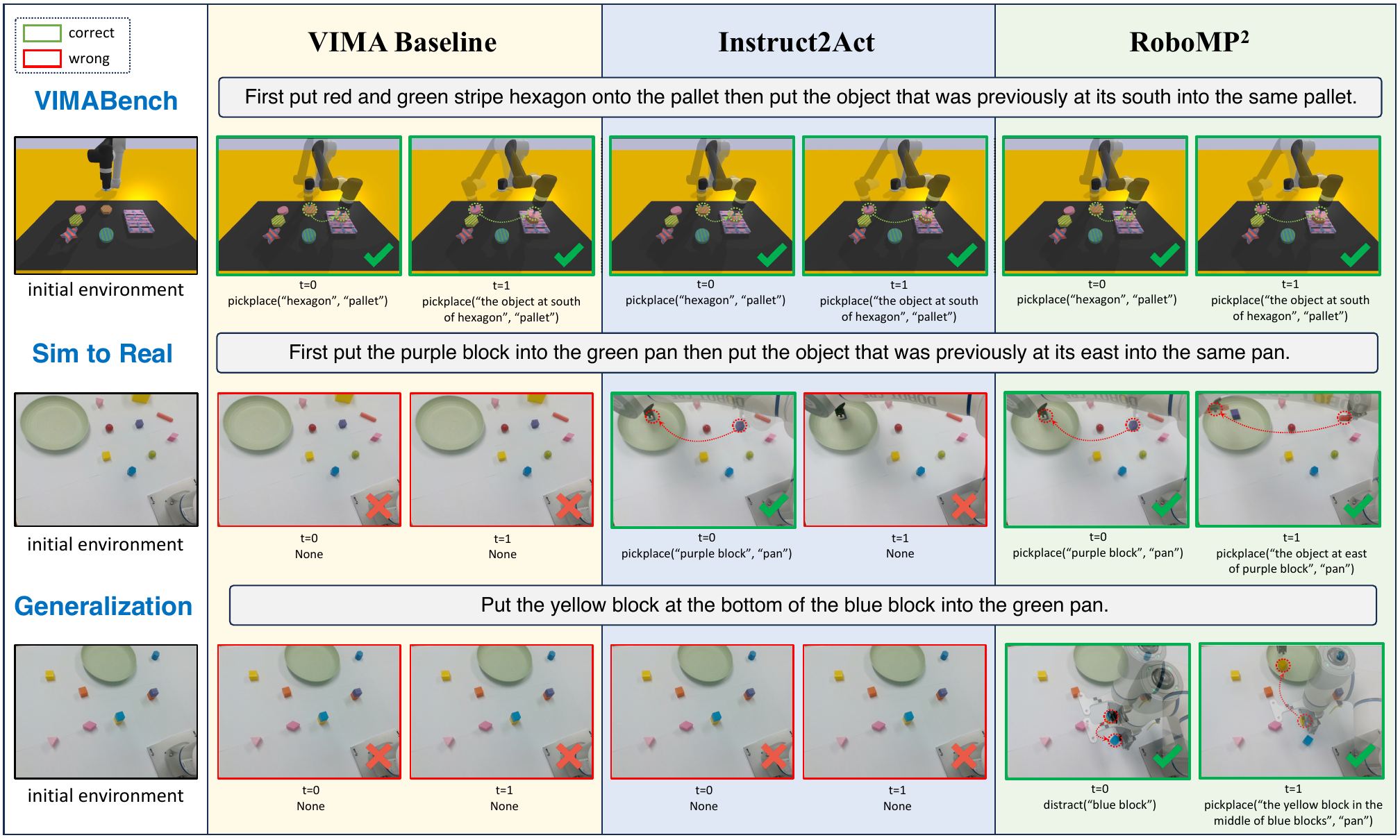}
    \caption{Visualization of robot planning and execution on VIMABench, zero-shot sim-to-real and generalization tasks.}
    \label{fig:visualization}
    \vspace{-8pt}
\end{figure*}

\subsection{Ablation Study}
\textbf{Effects of the Coarse-to-fine Retriever.}
We conduct ablation experiments to valid the performance of RoboMP$^2$ with different settings of the retriever-augmented module.
In particular, we compare 4 different implementations:
(1) \textit{w/o CR} removes the coarse retriever, directly using the rewriting module to order all programs in the codebase;
(2) \textit{w/o IR} removes the rewriting module, remaining the coarse retriever and semantic reorder;
(3) \textit{w/o SR} removes the reorder module, unsorting the retrieved demonstrations;
(4) \textit{w/o RAMP} removes the whole retriever, and takes $k$ programs from the codebase randomly as prompt examples.
As depicted in \textit{Table}~\ref{tab:retriever_abl_1}, without either of these components, the average performance drops significantly.
The reason is that when the context is composed of irrelevant examples, it is difficult for the planner to generate execution plans successfully.
\begin{table}[h!]
    \centering    
    \resizebox{\linewidth}{!}{
        \begin{tabular}{l|rrr|llll|l}
    \toprule
          & \multicolumn{1}{c}{CR} & \multicolumn{1}{c}{IR} & \multicolumn{1}{c|}{SR} & \multicolumn{1}{c}{L1}    & \multicolumn{1}{c}{L2}    & \multicolumn{1}{c}{L3}    & \multicolumn{1}{c|}{L4}    & \multicolumn{1}{c}{Avg.} \\
    \midrule
    \multicolumn{1}{l|}{RoboMP$^2$\ \ \ \ } &    \Checkmark   &    \Checkmark   &  \Checkmark     & \textbf{89.0} & 85.9  & \textbf{86.8} & \textbf{68.0} & \textbf{82.4} \\
     RoboMP$^2$\ $_\mathrm{w/o\ CR}$     &    \XSolidBrush   &  \Checkmark     &   \Checkmark    & 14.4  & 16.0  & 19.6  & \ \ 8.0   & 14.5  \\
    RoboMP$^2$\ $_\mathrm{w/o\ IR}$      &   \Checkmark    &    \XSolidBrush   &   \Checkmark    & 87.7  & \textbf{86.1} & 86.7  & 49.0  & 77.4  \\
    RoboMP$^2$\ $_\mathrm{w/o\ SR}$      &   \Checkmark    &   \Checkmark    &   \XSolidBrush    & 81.9  & 80.6  & 82.7  & 65.0  & 77.5  \\
    RoboMP$^2$\ $_\mathrm{w/o\ RAMP}$      &    \XSolidBrush   &   \XSolidBrush    & \XSolidBrush      & 10.5  & \ \ 9.5   & 11.6  & \ \ 3.0   & \ \ 8.7  \\
    \bottomrule
    \end{tabular}
    }
    \caption{The performance of RoboMP$^2$ variants with removing different retriever components on VIMABench.}
    \label{tab:retriever_abl_1}
    \vspace{-10pt}
\end{table}
\label{sec:exp_retreiver}

\begin{table}[h!]
    \centering
    \resizebox{.74\linewidth}{!}{
    \begin{tabular}{llllll}
    \toprule
     & \multicolumn{1}{l}{$\lambda$=0}     & \multicolumn{1}{l}{$\lambda$=0.25}   & \multicolumn{1}{l}{$\lambda$=0.5}   & \multicolumn{1}{l}{$\lambda$=0.75}  & \multicolumn{1}{l}{$\lambda$=1} \\
    \midrule
    TF-IDF & 74.1  & \textbf{82.4}  & 72.4  &   72.6    & 73.9 \\
    BM25  &  42.8 & 43.1    &  42.7     & 42.8      &  41.8     \\
    S-BERT &  35.9     &  33.9     &   35.7    & 35.9     & 34.2 \\
    \bottomrule
    \end{tabular}}
    \caption{The comparison results regarding different values of $\lambda$ and similarity metrics on the VIMABench dataset.}
    \vspace{-3pt}
\end{table}

In addition, $\lambda$ in RAMP balances the contributions between the similarity score of the coarse retriever and the fine reranker.
We investigate its effects with different values ranging from 0 to 1, as well as different similarity calculation methods for the coarse retriever, including TF-IDF, BM25, and SentenceBERT.
As illustrated in \textit{Table}~\ref{tab:retriever_abl_1}, it achieves the best performance when $\lambda$ is set to 0.25 and TF-IDF performs consistently better than the other similarity calculation methods for the coarse retreiver.
Since the input for the fine reranker is derived from candidates provided by the coarse retriever, the distinctions among these candidates are minimal. 
Consequently, it is essential to assign a higher weight to the fine reranker to ensure its significant impact during the reranking process.
On the other hand, instructions for the same task often contain representative keywords on VIMABench.
Therefore, the TF-IDF similarity calculation method which is based on keyword frequency matching could achieve excellent results.

To investigate the robustness of RoboMP$^2$, we construct an extensive test set by enriching the original instructions using GPT4. 
Subsequently, we conduct experiments to evaluate the impact of this augmentation.
As shown in \textit{Table}~\ref{tab:exp_enrich}, the results demonstrate the outstanding semantic understanding capability of our approach. 
In comparison to I2A, RoboMP$^2$ is not confined to processing text instructions with fixed patterns. 
On the contrary, it possesses stronger generalization and robustness.

\textbf{Effectiveness of Multimodal Planner.}
To assess the influence of multimodal information on plan generation, we replaced the planning generator from GPT4V, which accepts both text and images as the input, with GPT3.5 and GPT4, both of which rely solely on the text input. 
The results on the VIMA Benchmark are illustrated in \textit{Table}~\ref{tab:mm_abl}.
It can be seen that the performance partially decreases when the visual input is absent. 
Upon analyzing the generated code plans, we found GPT4 fails to detect these anomalies, when there are disturbances in the environment without explicit mention in the task, as illustrated in \textit{Figure}~\ref{fig:intro_1}(b).
In contrast, with the assistance of visual information, GPT4V can effectively perceive environmental factors, enabling it to make correct decisions. 
\begin{table}[h!]
    \centering
  \resizebox{\linewidth}{!}{
    \begin{tabular}{l|cc|cccc|r}
    \toprule
          & \multicolumn{1}{c}{Text} & \multicolumn{1}{c|}{Image} & L1    & L2    & L3    & L4    & Avg. \\
    \midrule
    RoboMP$^2\ _\mathrm{w/\ GPT4V}$ &  \Checkmark     &  \Checkmark     & \textbf{89.0}    & \textbf{85.9}    & \textbf{86.8}   & \textbf{68.0}    & \textbf{82.4} \\
    RoboMP$^2\ _\mathrm{w/\ GPT4}$  &  \Checkmark    &   \XSolidBrush    & 87.8    & 79.6    & 77.8    & 65.0    & 77.6 \\
    RoboMP$^2\ _\mathrm{w/\ GPT3.5}$ &  \Checkmark     &  \XSolidBrush     & 67.2    & 73.2    & 70.7    & 53.9    & 66.2 \\
    \bottomrule
    \end{tabular}
    }
    \captionof{table}{The performance of multimodal and text-only planners.}
    \label{tab:mm_abl}
    \vspace{-12pt}
\end{table}

\textbf{Comparison of Multimodal Perceptors.}
To validate the capabilities of GCMP for intricate relationship references, we construct a perception test set.
This set contains fundamental perception tasks that detect objects by names, and complex perception tasks that requires understanding complex referential expressions as illustrated in Section~\ref{sec:LCP_complex_refer}.
Since the traditional visual perceptors used in the robotic manipulation, such as YOLO, cannot take the referential expressions to perceive objects, we compare our embodied perceptor GCMP with SAM+CLIP in I2A, and the general MLLM, i.e., Shikra~\cite{Shikra}.
It can be seen from \textit{Table}~\ref{tab:mm_grouding} that our GCMP outperforms the two models on both simple and complex referential perception by a large margin.
\begin{table}[h!]
    \centering
    \vspace{-6pt}
  \resizebox{\linewidth}{!}{
    \begin{tabular}{lrrr|rrr}
    \toprule
          & \multicolumn{3}{c|}{Simple Perception} & \multicolumn{3}{c}{Complex Perception} \\
\cmidrule{2-7} & \multicolumn{1}{r}{mAP$_{0.3}$}         & \multicolumn{1}{r}{mAP$_{0.5}$} & \multicolumn{1}{r|}{mAP$_{0.75}$}  & \multicolumn{1}{r}{mAP$_{0.3}$}& \multicolumn{1}{r}{mAP$_{0.5}$} & \multicolumn{1}{r}{mAP$_{0.75}$} \\
    \midrule
    SAM$_\mathrm{H}$+CLIP$_\mathrm{H}$ & 78.9 & 74.1  & 64.1   & 0     & 0     & 0 \\
    Shikra &    63.6   &    18.2   &   0    &   0    &   0    & 0  \\
    GCMP  &\textbf{99.1} & \textbf{94.0}    & \textbf{89.2} & \textbf{99.0}&\textbf{97.7} & \textbf{78.0} \\
    \bottomrule
    \end{tabular}
    }
    \captionof{table}{The performance of multimodal perceptors.}
    \label{tab:mm_grouding}
\end{table}
Neither SAMP+CLIP or Shikra can work on the complex referential perception.
The reason is that SAM+CLIP does not have a ability to understand the complex semantic reference and the general model Shikra struggles to generalize to complex embodied perception tasks.

\subsection{Qualitative Results}
We further present qualitative results in \textit{Figure}~\ref{fig:visualization}, illustrating the planning and execution steps of RoboMP$^2$ in comparison to other methods across three different kinds of tasks.
It can be observed that only RoboMP$^2$ successfully accomplishes all tasks.
While all three methods perform well on VIMABench tasks, the end-to-end policy fails to generalize to real-world scenarios due to differences in the robot arm.
Furthermore, SAM+CLIP schema is shown to unable to recognize referential objects with complex relationships. This leads I2A to only complete pick-and-place tasks for simple objects, resulting in failure in the second step of sim-to-real. 
In the generalization scenario, other methods are unable to complete the instruction. 
Only RoboMP$^2$, leveraging its robust language understanding and multimodal perception capabilities, successfully identify objects with intricate relationships. 
It adaptively generates action strategies based on the current environment, ultimately completing the task.

\section{Conclusion}
In this paper, we have proposed a novel Robotic Perception and Planning framework (RoboMP$^2$) that consists of the Goal-Conditioned Multimodal Perceptor (GCMP) and the Retrieval-Augmented Multimodal Planner (RAMP).
GCMP is introduced to capture multimodal environment information by incorporating a tailored MLLM. 
RAMP employs a coarse-to-fine retrieval-augmented approach to adapatively select the $k$ most-relevant policies as in-context demonstrations to enhance the generalization.
Extensive experiments demonstrate that RoboMP$^2$ outperforms the baselines by a large margin on both VIMABench and real-world tasks.
\section*{Impact Statement}
Inspired by the great successes of multimodal large language models (MLLMs), we propose RoboMP$^2$ that leverages both multimodal information in an environment and the general intelligence in MLLMs to enhance the perception and reasoning capabilities of embodied robots.
To this end, this technology is expected to advance smart robots that can free human beings from some tedious work.
There are many potential societal consequences of developing smart robots, none which we feel must be specifically highlighted here.

\section*{Acknowledgements}
We would like to thank the reviewers for their constructive comments.
This work is supported by Shenzhen College Stability Support Plan (Grant No.GXWD20220817144428005).
Additionally, it is also supported by China Postdoctoral Science Foundation (Grant No. 2023TQ0095 and 2023M740923) and partially supported by National Natural Science Foundation of China (Grant No.62306090).

\bibliography{bib}
\bibliographystyle{icml2024}

\newpage
\appendix
\onecolumn

\section{Similarity Computation Method}\label{appendix:dense_retriever}
(1) \textbf{TF-IDF}: It is a numerical statistic to evaluate the importance of a word within a document relative to a collection of corpus.
We obtain the term frequency (TF) and the inverse document frequency (IDF), then multiplying them to compute the the TF-IDF score as followed:
\begin{gather}
    \mathrm{TF}(t_i, q) = \frac{n_{t_i, q}}{\sum_{k=1}^{l} n_{t_k, q}}, \\
    \mathrm{IDF}(t_i, \mathcal{C}_j) = \log\left({\frac{|{\mathcal{C}|}}{n_{t_i, \mathcal{C}_j}+ 1}}\right),  \\
    \mathrm{Score}_{\mathrm{TF-IDF}}(q, \mathcal{C}_j) = \sum \mathrm{IDF}(t_i, \mathcal{C}_j) \times \mathrm{TF}(t_i, q),
\end{gather}
where $t_i$, $q$, $\mathcal{C}_j$ represents the term word, query, and program in codebase $\mathcal{C}$.
$n_{t_i,q}$ means the number of times term $t_i$ appears in query $q$ while $|\mathcal{C}|$ denotes its total number.

(2) \textbf{BM25}: This is an extension of TF-IDF, particularly effective in dealing with long documents. It improves the calculation of IDF and simultaneously enhances the weighting of TF, instead of the original TF.
The computation procedure is as followed:
\begin{gather}
    \mathcal{W}(t_i, q, \mathcal{C}) = \frac{\mathrm{TF}(t_i, q)\cdot (k_1+1)}{\mathrm{TF}(t_i, q)+k_1\cdot (1-b+b\cdot \frac{|q|}{L_{avg}({\mathcal{C}})})}, \\
    \mathrm{IDF}(t_i, \mathcal{C}_j) = \log(\frac{|\mathcal{C}|-n_{t_i, \mathcal{C}_j}+0.5}{n_{t_i, \mathcal{C}_j} + 0.5} + 1), \\
    \mathrm{Score}_{\mathrm{BM25}}(q, \mathcal{C}_j)=\sum \mathrm{IDF}(t_i, \mathcal{C}_j)\cdot \mathcal{W}(t_i, q),
\end{gather}
where $\mathcal{W}$ means the weight of each term $t_i$ in query $q$, and $L_{\mathrm{avg}}$ denotes the average length of codebase $\mathcal{C}$.

(3) \textbf{SentenceBERT}:
Different from keyword similarity computation methods, it embeds the sentence to a vector via a pretrain language model, and then use the cosine similarity to compute their scores.
\begin{gather}
    h_q = \textrm{Encoder}(q) \\ 
    h_{\mathcal{C}_j} = \textrm{Encoder}(\mathcal{C}_j) \\ 
    \mathrm{Score}_{\mathrm{SBERT}} = h_q \circ h_{\mathcal{C}_j}
\end{gather}

\section{Rewriting Prompt}\label{appendix:rewriting_prompt}
\begin{figure}[htbp]
    \centering
    \includegraphics[width=\linewidth]{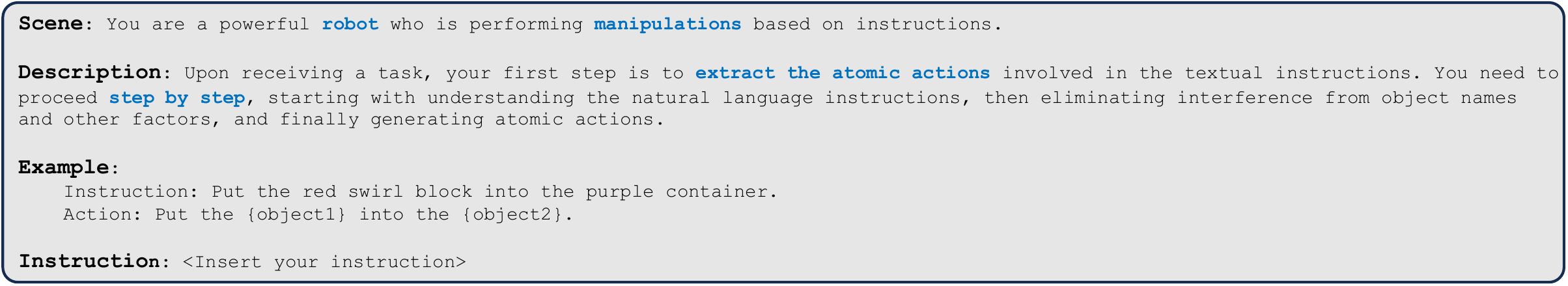}
    \caption{The soft template of instruction rewritter.}
    \label{fig:soft_template}
\end{figure}

As shown in \textit{Figure}~\ref{fig:soft_template}, we construct a rewritter soft template which consists of robotic scene, description, example and instruction.
The first two primarily provide premises and requirements for the robot scenarios, while the latter two consist of concrete examples and the corresponding task instruction.

\section{Dataset}\label{appendix:dataset}
\subsection{VIMABench}
VIMABench (Jiang et al., 2022) introduces a comprehensive four-level evaluation protocol designed to progressively increase in difficulty, challenging trained agents. The protocol comprises four levels, each testing different aspects of the agent's capabilities.
\begin{itemize}
    \item Level 1 (L1) - Placement Generalization: Randomizing the placement of target objects is required in this level, testing the agent's ability to adapt to varying configurations.
    \item Level 2 (L2) - Combinatorial Generalization: Generating new combinations of target materials and object descriptions challenges the agent's adaptability to diverse scenarios.
    \item Level 3 (L3) - Novel Object Generalization: Assessing the agent’s capacity to generalize to unfamiliar materials and objects, gauging its ability to handle the unknown.
    \item Level 4 (L4) - Novel Task Generalization: Requiring agents to ground and execute tasks that have not been encountered before, pushing the boundaries of the agent's problem-solving capabilities.
\end{itemize}

The tasks, presented in the order of their corresponding task index, align with the structure of VIMABench:
\begin{enumerate}[itemsep=1pt,topsep=0pt,parsep=0pt]
    \setlength{\itemsep}{2pt}
    \setlength{\parsep}{2pt}
    \setlength{\parskip}{1pt}
    \item Visual Manipulation: Select a designed object and place it into a specific container.
    \item Scene Understanding: Recognize a specific object in a provided image and place it into a designed container.
    \item Rotate: Rotate a specific object by a designated degree.
    \item Rearrange: Target objects in the current state into the scene pattern based on a provided image.
    \item Rearrange then Restore: Complete the Rearrange task, then restore objects to their initial placement.
    \item Novel Adjectives Understanding: Recognize novel adjectives by comparing object images provided in the VIMABench prompt.
    \item Novel Nouns: Recognize novel nouns similar to the above.
    \item Novel Nouns and Adjectives: Combine novel adjectives and nouns effectively.
    \item Novel Concept Understanding: Recognize degrees indicated in the prompt and rotate a specific object accordingly.
    \item Follow Motion: Manipulate objects corresponding to given frames in the VIMABench prompt.
    \item Stack the Blocks: Similar to the above, involving stacking blocks.
    \item Sweep Without Exceeding: Manipulate specific objects without exceeding a constraint.
    \item Sweep Without Touching: Manipulate specific objects without touching a constraint.
    \item Same Texture: Recognize objects with the same texture and manipulate them into a specific container.
    \item Same Shape: Recognize objects with the same shape and manipulate them into a specific container.
    \item Manipulate the Old Neighbor: Place an object into a specific container, then place an object in the direction of its original location into the same container.
    \item Pick in Order then Restore: Manipulate one object into different containers sequentially, and finally restore it to the initial container.
\end{enumerate}

\subsection{Real-world Tasks}
In the real-world setting, a robotic arm with a gripper is utilized. A camera is positioned parallel to the robotic arm for optimal viewing. Real task experiments focus on object localization, object attribute recognition, complex scenes, complex reasoning, and contextual memory.

Additional tasks in the real-world setting include:
\begin{enumerate}[itemsep=1pt,topsep=0pt,parsep=0pt]
    \setlength{\itemsep}{2pt}
    \setlength{\parsep}{2pt}
    \setlength{\parskip}{1pt}
    \item Visual Manipulation: It is the same as the VIMABench.
    \item Same Shape: It is the same as the VIMABench.
    \item Same Color: Recognize objects with the same color and manipulate them into a specific container.
    \item Manipulate Old Neighbor: It is the same as the VIMABench.
    \item Interfering Manipulation: There is a disturbance object in the environment and the robot need recognize the state. First distract the interfered object, then execute the following instruction.
\end{enumerate}

\begin{figure}[h!]
    \centering
    \includegraphics[width=0.85\linewidth]{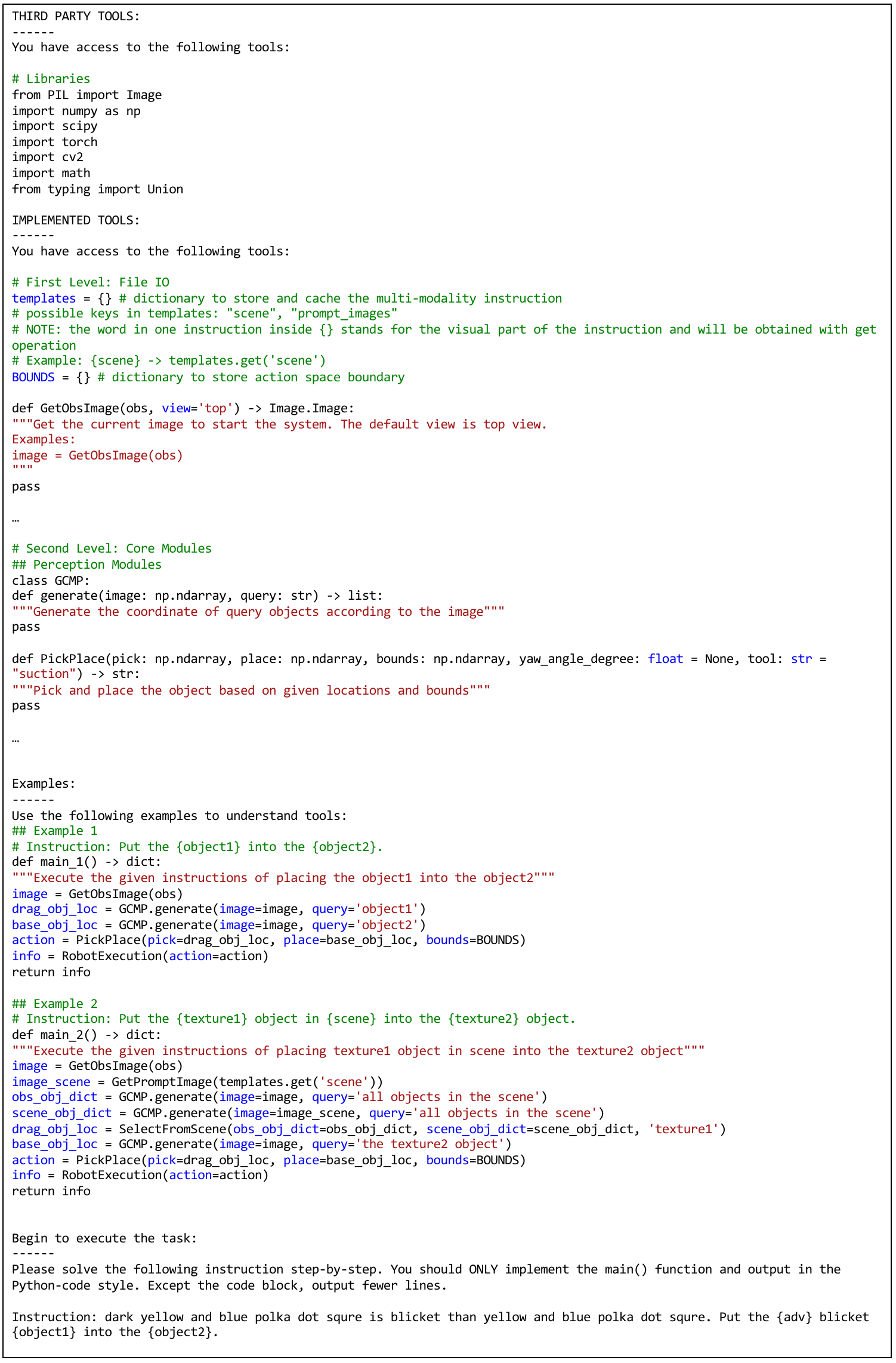}
    \caption{An example of complete template for the prompt multimodal generator.}
    \label{appendix:full_template}
\end{figure}
\section{Planning Template}\label{full_template}
We construct a template to prompt multimodal generator to generate planning.
An example of complete prompt is shown in Figure~\ref{appendix:full_template}.
By combining the specific in-context sample, the length of the whole input tokens would be approximately 8k.

\section{Supplementary Result}\label{appendix:full_result}
The VIMABench contains 17 tasks of L1-L4 levels.
We present the detailed experimental result in Table~\ref{tab:supplement_result}.
\begin{table}[!h]
  \centering
  \resizebox{\linewidth}{!}{
      \begin{tabular}{lrrrrrrrrrrrrrrrrrr}
    \toprule
    Method & 01    & 02    & 03    & 04    & 05    & 06    & 07    & 08    & 09    & 10    & 11    & 12    & 13    & 14    & 15    & 16    & 17    & Avg \\
    \midrule
    \rowcolor[rgb]{ 1,  .949,  .8} \multicolumn{19}{c}{\textbf{L1 Level}} \\
    \midrule
    \textbf{Gato} & 79.0  & 68.0  & 91.5  & 57.0  & 44.5  & 54.0  & 74.0  & -     & 18.0  & -     & 61.0  & 88.5  & -     & -     & 83.5  & 33.5  & 2.5   & 58.1  \\
    \textbf{Flamingo} & 56.0  & 58.5  & 63.0  & 48.5  & 38.0  & 48.5  & 62.5  & -     & 3.5   & -     & 66.5  & 86.0  & -     & -     & 40.0  & 43.5  & 2.5   & 47.5  \\
    \textbf{GPT} & 62.0  & 57.5  & 41.0  & 55.5  & 45.5  & 47.5  & 54.5  & -     & 8.5   & -     & 77.0  & 81.5  & -     & -     & 41.0  & 38.0  & 0.5   & 46.9  \\
    \textbf{VIMA} & 100.0  & 100.0  & 99.5  & 100.0  & 56.5  & 100.0  & 100.0  & -     & 18.0  & -     & 77.0  & 93.0  & -     & -     & 97.0  & 76.5  & 43.0  & 81.6  \\
    \textbf{RT-2} & 100.0  & 98.0  & 97.0  & 58.0  & 30.0  & 98.0  & 97.0  &       & 14.0  &       & 84.0  & 79.0  &       &       & 93.0  & 52.0  & 47.0  & 72.8  \\
    \midrule
    \textbf{CaP} & 90.0  & 80.8  & 96.0  & 65.3  & 61.3  & 80.0  & 84.7  & -     & 38.0  & -     & 68.0  & 60.0  & -     & -     & 67.3  & 58.0  & 76.0  & 71.2  \\
    \textbf{VisualProg} & 92.0  & 27.0  & 9.0   & 29.0  & 90.0  & 38.0  & 87.0  & -     & 21.3  & -     & 65.3  & 30.7  & -     & -     & 92.0  & 36.7  & 28.7  & 49.7  \\
    \textbf{I2A} & 91.3  & 81.4  & 98.2  & 78.5  & 72.0  & 82.0  & 88.0  & -     & 42.0  & -     & 72.0  & 68.0  & -     & -     & 78.0  & 64.0  & 85.2  & 77.0  \\
    \rowcolor[rgb]{ .906,  .902,  .902} \textbf{RoboMP$^2$} & 100.0  & 89.3  & 100.0  & 92.7  & 95.3  & 86.0  & 96.0  & -     & 55.3  &       & 89.3  & 71.3  & -     & -     & 86.0  & 98.7  & 96.7  & \textbf{89.0} \\
    \midrule
    \rowcolor[rgb]{ 1,  .949,  .8} \multicolumn{19}{c}{\textbf{L2 Level}} \\
    \midrule
    \textbf{Gato} & 56.5  & 53.5  & 88.0  & 55.5  & 43.5  & 55.5  & 53.0  & -     & 14.0  & -     & 63.0  & 90.5  & -     & -     & 81.5  & 33.0  & 4.0   & 53.2  \\
    \textbf{Flamingo} & 51.0  & 52.5  & 61.5  & 49.5  & 38.5  & 47.5  & 55.5  & -     & 5.5   & -     & 70.5  & 82.0  & -     & -     & 42.0  & 39.0  & 3.0   & 46.0  \\
    \textbf{GPT} & 52.0  & 52.0  & 49.5  & 54.5  & 45.5  & 52.5  & 51.0  & -     & 11.0  & -     & 76.5  & 84.0  & -     & -     & 43.0  & 38.0  & 0.5   & 46.9  \\
    \textbf{VIMA} & 100.0  & 100.0  & 99.5  & 100.0  & 54.5  & 100.0  & 100.0  & -     & 17.5  & -     & 77.0  & 93.0  & -     & -     & 98.5  & 75.0  & 45.0  & 81.5  \\
    \textbf{RT-2} & 100.0  & 96.0  & 97.0  & 56.0  & 27.0  & 95.0  & 97.0  &       & 10.0  &       & 84.0  & 83.0  &       &       & 92.0  & 43.0  & 34.0  & 70.3  \\
    \textbf{CaP} & 90.0  & 79.3  & 96.0  & 64.7  & 60.0  & 74.6  & 85.3  & -     & 37.3  & -     & 66.7  & 62.7  & -     & -     & 66.0  & 52.7  & 74.7  & 70.0  \\
    \textbf{VisualProg} & 84.0  & 26.0  & 11.3  & 38.7  & 87.3  & 30.0  & 80.7  & -     & 20.0  & -     & 71.3  & 22.0  & -     & -     & 94.7  & 24.0  & 30.0  & 47.7  \\
    \textbf{I2A} & 91.5  & 80.8  & 97.8  & 74.9  & 69.5  & 81.0  & 86.0  & -     & 44.0  & -     & 70.5  & 65.0  & -     & -     & 80.0  & 66.0  & 84.0  & 76.2  \\
    \rowcolor[rgb]{ .906,  .902,  .902} \textbf{RoboMP$^2$} & 99.3  & 78.7  & 100.0  & 91.3  & 91.3  & 82.0  & 92.0  & -     & 36.0  &       & 92.0  & 68.7  & -     & -     & 91.3  & 96.7  & 96.7  & \textbf{85.9} \\
    \midrule
    \rowcolor[rgb]{ 1,  .949,  .8} \multicolumn{19}{c}{\textbf{L3 Level}} \\
    \midrule
    \textbf{Gato} & 51.0  & 58.0  & 84.5  & 56.5  & 35.5  & 53.5  & 49.0  & -     & 15.0  & -     & 65.0  & -     & -     & -     & 52.0  & 33.0  & 0.0   & 43.5  \\
    \textbf{Flamingo} & 49.0  & 50.0  & 66.5  & 47.0  & 35.0  & 47.5  & 50.0  & -     & 4.0   & -     & 66.0  & -     & -     & -     & 30.5  & 43.5  & 0.5   & 40.8  \\
    \textbf{GPT} & 52.0  & 51.0  & 55.0  & 49.5  & 40.0  & 46.0  & 50.5  & -     & 5.0   & -     & 82.0  & -     & -     & -     & 37.0  & 38.0  & 1.5   & 42.3  \\
    \textbf{VIMA} & 99.0  & 100.0  & 100.0  & 97.0  & 58.0  & 100.0  & 99.0  & -     & 17.5  & -     & 90.5  & -     & -     & -     & 97.5  & 46.0  & 43.5  & 79.0  \\
    \textbf{RT-2} & 96.0  & 94.0  & 96.0  & 52.0  & 31.0  & 95.0  & 93.0  &       & 11.0  &       & 97.0  &       &       &       & 93.0  & 40.0  & 3.0   & 66.8  \\
    \textbf{CaP} & 90.0  & 79.3  & 95.3  & 63.3  & 60.0  & 74.0  & 84.7  & -     & 37.3  & -     & 66.0  & -     & -     & -     & 64.3  & 51.3  & 72.0  & 69.8  \\
    \textbf{VisualProg} & 83.3  & 25.3  & 13.3  & 27.3  & 62.0  & 32.0  & 80.7  & -     & 17.3  & -     & 70.0  & -     & -     & -     & 73.3  & 24.0  & 39.7  & 45.7  \\
    \textbf{I2A} & 91.8  & 80.2  & 97.4  & 81.8  & 65.8  & 79.0  & 89.0  & -     & 38.0  & -     & 71.0  & -     & -     & -     & 78.0  & 62.0  & 82.0  & 76.3  \\
    \rowcolor[rgb]{ .906,  .902,  .902} \textbf{RoboMP$^2$} & 100.0  & 83.3  & 100.0  & 86.0  & 86.0  & 78.7  & 95.3  &   -    & 40.7  &  -     & 92.0  &   -    &     -  &  -     & 89.3  & 98.0  & 92.7  & \textbf{86.8} \\
    \midrule
    \rowcolor[rgb]{ 1,  .949,  .8} \multicolumn{19}{c}{\textbf{L4 Level}} \\
    \midrule
    \textbf{Gato} & -     & -     & -     & -     & -     & -     & -     & 20.5  & -     & 0.0   & -     & -     & 0.0   & 29.0  & -     & -     & -     & 12.4  \\
    \textbf{Flamingo} & -     & -     & -     & -     & -     & -     & -     & 21.0  & -     & 0.0   & -     & -     & 0.0   & 27.5  & -     & -     & -     & 12.1  \\
    \textbf{GPT} & -     & -     & -     & -     & -     & -     & -     & 20.5  & -     & 0.5   & -     & -     & 0.0   & 36.0  & -     & -     & -     & 14.3  \\
    \textbf{VIMA} & -     & -     & -     & -     & -     & -     & -     & 100.0  & -     & 0.0   & -     & -     & 0.0   & 95.5  & -     & -     & -     & 48.9  \\
    \textbf{RT-2} & -     & -     & -     & -     & -     & -     & -     & 98.0  & -     & 0.0   & -     & -     & 0.0   & 90.0  & -     & -     & -     & 47.0  \\
    \textbf{CaP} & -     & -     & -     & -     & -     & -     & -     & 74.0  & -     & 28.0  & -     & -     & 0.0   & 76.7  & -     & -     & -     & 44.7  \\
    \textbf{VisualProg} & -     & -     & -     & -     & -     & -     & -     & 52.0  & -     & 64.0  & -     & -     & 0.0   & 92.7  & -     & -     & -     & 52.2  \\
    \textbf{I2A} & -     & -     & -     & -     & -     & -     & -     & 84.0  & -     & 35.0  & -     & -     & 0.0   & 80.0  & -     & -     & -     & 49.8  \\
    \rowcolor[rgb]{ .906,  .902,  .902} \textbf{RoboMP$^2$} & -     & -     & -     & -     & -     & -     & -     & 76.7  & -     & 98.7  & -     & -     & 0.0   & 96.7  & -     & -     & -     & \textbf{68.0} \\
    \bottomrule
    \end{tabular}%
    }%
  \caption{The full result of all tasks on VIMABench. ``-'' denotes that the task is excluded from this level.}
  \label{tab:supplement_result}%
\end{table}%

\end{document}